\documentclass[conference]{IEEEtran}
\IEEEoverridecommandlockouts

\usepackage{cite}
\usepackage{amsmath,amssymb,amsfonts}
\usepackage{listings}
\usepackage{algorithmic}
\usepackage{graphicx}
\usepackage{textcomp}
\usepackage{xcolor}
\usepackage{float}
\usepackage{multirow}
\usepackage{xurl}

\IEEEoverridecommandlockouts

\begin{document}

\title{Going NUTS with ADVI: Exploring various Bayesian inference techniques with Facebook Prophet\thanks{This paper was partially supported by Faculty of Computer Science and Engineering, Skopje, Macedonia}}

\author{\IEEEauthorblockN{Jovan Krajevski}
\IEEEauthorblockA{\textit{Faculty of Computer Science and Engineering} \\
\textit{Ss. Cyril and Methodius University}\\
Skopje, Macedonia \\
jovan.krajevski@gmail.com}
\and
\IEEEauthorblockN{Biljana Tojtovska Ribarski}
\IEEEauthorblockA{\textit{
Faculty of Computer Science and Engineering} \\
\textit{Ss. Cyril and Methodius University}\\
Skopje, Macedonia \\
biljana.tojtovska@finki.ukim.mk}
}

\maketitle

\begin{abstract}
Since its introduction, Facebook Prophet has attracted positive attention from both classical statisticians and the Bayesian statistics community. The model provides two built-in inference methods: maximum a posteriori estimation using the L-BFGS-B algorithm, and Markov Chain Monte Carlo (MCMC) sampling via the No-U-Turn Sampler (NUTS). While exploring various time-series forecasting problems using Bayesian inference with Prophet, we encountered limitations stemming from the inability to apply alternative inference techniques beyond those provided by default. Additionally, the fluent API design of Facebook Prophet proved insufficiently flexible for implementing our custom modeling ideas. To address these shortcomings, we developed a complete reimplementation of the Prophet model in PyMC, which enables us to extend the base model and evaluate and compare multiple Bayesian inference methods. In this paper, we present our PyMC-based implementation and analyze in detail the implementation of different Bayesian inference techniques. We consider full MCMC techniques, MAP estimation and Variational inference techniques on a time-series forecasting problem. We discuss in details the sampling approach, convergence diagnostics, forecasting metrics as well as their computational efficiency and detect possible issues which will be addressed in our future work. 
\end{abstract}


\section{Introduction}

Performing full Bayesian inference has always posed significant computational challenges. While the statistical quality obtained by sampling from the true posterior using Markov Chain Monte Carlo (MCMC) methods is widely recognized and rarely questioned, the computational cost involved often raises doubts about whether it is worthwhile to perform sampling. On the other hand, maximum a posteriori (MAP) estimates are computationally efficient but they collapse the full posterior distribution into a single point estimate, discarding information about parameter uncertainty that is central to Bayesian analysis.

Variational Inference (VI) is a middle-ground approach that bridges the gap between computationally demanding MCMC sampling and overly simplistic MAP estimation. By providing an analytical approximation to the true posterior and sampling from it, VI significantly reduces computational overhead compared to full MCMC sampling. However, like many other approximation techniques, it does not always capture all essential characteristics of the posterior.

There is no universally optimal Bayesian inference method - the choice of an optimal method depends on the model complexity, data size and inferential goals. The goal of this paper is to study the forecasting performance and computational accuracy of different Bayesian inference methods. We analyze the application of full MCMC sampling, VI, and MAP estimation and their various algorithmic implementations on a specific model and forecasting scenario. This investigation arose during our exploration of Bayesian transfer learning techniques for time-series forecasting tasks using Facebook Prophet \cite{taylor2018forecasting}. Throughout our research, we quickly encountered limitations related to Prophet's fluent API design \cite{fowler2005fluent}, which restricted our ability to incorporate new components into Prophet's decomposable time-series model \cite{harvey1990estimation}. Additionally, Prophet's built-in inference methods proved insufficient: full MCMC sampling was computationally prohibitive for rapid experimentation, while MAP estimation alone did not adequately represent parameter uncertainty. To overcome these limitations, we developed a complete reimplementation of Facebook Prophet using PyMC \cite{abril2023pymc}.  

In this paper, we first go over our PyMC-based reimplementation. We give a quick overview of the FP model, before introducing the new intuitive API that we will implement. Section II also includes  theoretical introduction of various Bayesian inference techniques. The methodology for their comparison and the results are presented in Section III and IV. The conclusions and future work  are given in Section V and VI. 

\section{Reimplementing Facebook Prophet in PyMC}

\subsection{Overview of the model}

Facebook Prophet is a decomposable time series model given by

\begin{equation}\label{fbprophet_def}
    y(t) \sim g(t) + \sum_{i=1}^{M}s(t;p_i,c_i) + h(t) + \epsilon(t)
\end{equation}
where \(y(t)\) is the time-series that is being modeled, \(g(t)\) is a trend component, \(s(t;p_i,c_i)\) is a Fourier seasonality component with a period of \(p_i\) represented with a series of order \(c_i\), \(h(t)\) is a holiday component and $\epsilon(t)$ is normally distributed random error \footnote{This case is known as "additive seasonality". If the seasonality is multiplicative, (\ref{fbprophet_def}) becomes
\(y(t) \sim g(t)\Big(1 + \sum_{i=1}^{M}s(t;p_i,c_i) + h(t)\Big) + \epsilon(t)\).
}.

Facebook Prophet allows the user to specify additional regressors. However, in this paper and in our future research, we work with univariate time-series, so additional regressors will be ignored.

Facebook Prophet gives an option of three different choices for the trend component $g(t)$: a piece-wise linear trend, a piece-wise logistic trend and a flat trend. In the piece-wise trends, it is up to the user to define the number of change points that will be taken into consideration. Each new change point introduces a new parameter that needs to be estimated, increasing the computational complexity of the model. By default, Facebook Prophet uses 25 change points.

The user is allowed to specify the series order of the Fourier seasonality component $s(t;p_i,c_i)$, i.e. the number of sine and cosine terms that will be used to approximate the periodical seasonality. Increasing the series order by one introduces one additional sine and one additional cosine term, by which  the number of parameters that need to be estimated is increased by two. By default, Facebook Prophet fits two seasonality components: a yearly seasonality (period of 365.25) approximated with a Fourier series of order 10, and a weekly seasonality (period of 7) approximated with a Fourier series of order 3.

In the rest of this paper, the holiday component will be removed since it is not significant for our problem. 

\subsection{Intuitive API}

Facebook Prophet's current implementation lacks the flexibility needed to introduce new components and only allows the user to do NUTS sampling or perform a MAP estimate. Its API employs a fluent interface, which is also less intuitive than the mathematical notation used to specify the model.

In our reimplementation of Facebook Prophet, we guide the development of the API by following other reimplementations in PyMC \cite{vink, pmprophet2020, airpassengers}. The resulting API, inspired by TimeSeers \cite{timeseers2021}, mirrors the mathematical notation typically used when writing models.

The main idea behind this API is to be modular - the user is allowed to define each component separately. So, if the user wishes to define a linear trend component with 10 changepoints, they would do that by instantiating an object from a \verb|LinearTrend| component by simply writing \verb|trend = LinearTrend(n_changepoints=10)|.

After defining the components, constructing the model is straightforward, because the API implements the summation and multiplication operators for these objects. If the user defined a linear trend \verb|trend|, a yearly seasonality \verb|yearly| and a weekly seasonality \verb|weekly|, they can create a multiplicative seasonality model by simply writing \verb|model = trend * (1 + yearly + weekly)|.

Below, we present the same Prophet model expressed in the fluent API from Facebook Prophet and in the more intuitive API we used in our reimplementation.

\begin{lstlisting}[basicstyle=\small,language=Python,caption=Two different APIs for specifying Prophet-like models]

#fluent API from Facebook Prophet
model = Prophet(
    seasonality_mode="multiplicative",
    yearly_seasonality=10,
    weekly_seasonality=3,
).add_seasonality(
    name="monthly", 
    period=30.5, 
    fourier_order=5,
)

#new intuitive API
model = LinearTrend() * (
    1
    + FourierSeasonality(p=365.25, order=10)
    + FourierSeasonality(p=30.5, order=5)
    + FourierSeasonality(p=7, order=3)
)
\end{lstlisting}

After introducing the new API, we briefly go over the details of full MCMC sampling, MAP estimation and VI.

\subsection{Full MCMC sampling}

We investigate three different algorithms that perform MCMC sampling: Metropolis-Hastings (MH)\cite{metropolis1953equation, hastings1970monte}, NUTS\cite{hoffman2014no} and Z-score adaptation of DEMetropolis (DMZ)\cite{ter2008differential}. These algorithms differ essentially. MH is the original random-walk approach in sampling from the posterior. The NUTS algorithm is an extension of the Hamiltonian  Monte Carlo (HMC) - it requires computation of the gradient of the log posterior, which points to the high-density regions of the posterior. On the other hand DMZ is a population-based, gradient-free MCMC algorithm that uses differential evolution between chains in a standardized Z-space as its proposal mechanism.

All of these approaches generate samples that approximate the true posterior distribution. However, due to limitations in the sampling algorithms and the inherent stochasticity of MCMC, we typically require a large number of samples to obtain reliable statistical estimates. MCMC sampling usually includes a warm-up (or adaptation) phase, during which the algorithm adjusts its internal parameters and moves away from the initial starting point toward areas of higher posterior density. This process is typically performed across multiple independent chains. Warm-up samples are generally discarded before inference begins. 

\subsection{MAP estimate}

The MAP estimate equals the mode of the posterior distribution. It is Bayesian regularized generalization of maximum likelihood estimation (MLE) - the maximum likelihood function is modified by including a prior distribution over the parameters, effectively regularizing the solution according to prior knowledge. Since this is a point estimate, it fails to capture all of the characteristics of the posterior and closely resembles the philosophy behind frequentist inference. For example, given a multi-modal posterior, the MAP estimate will not inform us of the existence of the other "modes". 

In this paper, we will investigate performin MAP using L-BFGS-B\cite{byrd1995limited}. It is a fast, gradient based (quasi-Newton) optimization algorithm which is memory efficient in presence of large number of parameters and thus suitable for MAP estimation.

\subsection{Variational Inference}

In VI, the true posterior is approximated by a variational distribution with a known analytical form, usually by minimizing the Kullback–Leibler divergence between the true posterior and the variational distribution. The sampling is performed from the variational distribution, which is a faster process than sampling from the true posterior.

In this paper, we investigate two different VI algorithms: ADVI\cite{kucukelbir2017automatic} and FullRank ADVI (FR ADVI)\cite{Blei_2017} ADVI employs a factorized Gaussian approximation for computational efficiency, while FR ADVI enhances this by modeling parameter correlations through a full-rank Gaussian.  For more details we refer to the existing literature.

\section{Methodology}

\subsection{Dataset}

For analysis of the introduced techniques we choose a time series of the log daily page views for the Wikipedia page of Peyton Manning, available at the Facebook Prophet documentation and GitHub repository. We use two years of data to train the model and we forecast the next year.

\subsection{Model specification}

We use the default model offered by Facebook Prophet, without doing any hyperparameter tuning. This model uses a piecewise linear trend with 25 change points and additive seasonality that takes into account yearly and weekly seasonal patterns.

In Table \ref{tab0}, we present the parameters that we are sampling from. For each parameter, we denote the shape i.e. the number of independent random variables with the same prior.

\begin{table}[H]
    \caption{Parameters that are being sampled in the Facebook Prophet model}
    \begin{center}
        \begin{tabular}{|c|c|c|c|}
        \hline
        \textbf{Component} & \textbf{Parameter} & \textbf{Shape} & \textbf{Prior} \\
        \hline
        {Linear} & slope - \(k\) & 1 & \(Normal(0, 5)\)\\ 
        \cline{2-4}
        Trend - \(g(t)\) & intercept - \(m\) & 1 & \(Normal(0, 5)\)\\ 
        \cline{2-4}
        & change points - \(\delta\) & 25 & \(Laplace(0, 0.05)\)\\ 
        \hline
        Yearly & Fourier  & \(2 \cdot 10\) & \(Normal(0, 10)\) \\
        Seasonality - & coefficients - \(\beta_y\) & & \\
        \(s(t;365.25, 10)\) & & & \\
        \hline
        Weekly & Fourier  & \(2 \cdot 3\) & \(Normal(0, 10)\) \\
        Seasonality - & coefficients - \(\beta_w\) & & \\
        \(s(t;7, 3)\) & & & \\
        \hline
        \end{tabular}
    \label{tab0}
    \end{center}
\end{table}

\subsection{Sampling, Convergence Diagnostics and Forecasting Metrics}

When using the NUTS method, we drew 4 chains of 2000 samples each with a warm-up of 1000 samples that we discarded. We ran MH and DMZ with 1000000 samples, because the Effective Sample Size was too low, whereas the R-hat Diagnostic was too high for a smaller amount of samples. The warm-up for MH and DMZ consisted of 10000 samples that were discarded. For the VI techniques, we produced 2000 samples after running 100000 iterations. 

We additionally included convergence diagnostics. For the full MCMC sampling, we analyzed and present the plot of the posterior samples, the autocorrelation plot of the posterior samples, the Effective Sample Size (ESS) and the R-hat (Gelman-Rubin) diagnostics. For the ADVI and FR ADVI techniques, we analyzed and present the negative Evidence Lower Bound (ELBO) convergence. 

For each of the different inference methods, we calculated and compared the standard forecasting metrics (MAE, MSE, RMSE, MAPE) and the execution time. For those methods that reached convergence, we also present their forecasting metrics and execution time at the moment they reached convergence.

The simulations were executed on a machine with AMD 7950x processor with 64GB DDR5 system RAM running Debian 12.

\section{Results}

\subsection{MCMC convergence}
\subsubsection{R-hat}
The R-hat diagnostic statistic compares the between-chain and in-chain variances. R-hat value larger than 1.01 is evidence of a lack of convergence of the sampling algorithm. In Table \ref{tab_rhat}, where the R-hat diagnostic is calculated for each parameter from Table \ref{tab0}, we can see that only the NUTS sampling algorithm produces chains that converge.

\begin{table}[H]
    \caption{Comparing the R-hat diagnostic for each parameter}
    \begin{center}
        \begin{tabular}{|c|c|c|c|c|c|}
        \hline
        \textbf{Method} & \(k\) & \(m\) & \(\delta\) & \(\beta_y\) & \(\beta_w\) \\
        \hline
        \textit{MH} & 1.0443 & 1.0249 & 1.0032 & 1.0010 & 1.0000\\
 \hline
\textit{DMZ} & 1.4047 & 1.4168 & 1.5264 & 1.1215 & 1.0205\\
 \hline
\textit{NUTS} & 1.0005 & 1.0010 & 1.0010 & 1.0004 & 1.0004\\
 \hline
        \end{tabular}
    \label{tab_rhat}
    \end{center}
\end{table}

\subsubsection{Efective Sample Size}
The ESS for an MCMC method quantifies the number of “independent” samples in the chain, taking into account the autocorrelation between successive samples. Higher ESS implies that the estimates based on the chain (as means, variances etc.) are more reliable. We adopt $ESS > 400$ as the threshold for sufficient independent samples to ensure stable posterior estimates.

Drawing 1000000 samples with MH and DMZ did not yield an ESS larger than 400 for each parameter. The values are presented in Table \ref{tab_ess}. The small number of independent samples suggests that the chains are not exploring the posterior distribution efficiently. ESS values for NUTS for all parameters are significantly higher and this was achieved after drawing 2000 samples. 

\begin{table}[H]
    \caption{Comparing the ESS for each parameter}
    \begin{center}
        \begin{tabular}{|c|c|c|c|c|c|c|}
        \hline
        \textbf{Method} & \(k\) & \(m\) & \(\delta\) & \(\beta_y\) & \(\beta_w\)\\
        \hline
  \textit{MH} & 82.78 & 163.80 & 5697.00 & 274152.10 & 795493.71\\
 \hline
\textit{DMZ} & 8.66 & 8.52 & 9.70 & 86.80 & 328.91\\
 \hline
\textit{NUTS} & 3050.72 & 3267.21 & 7033.49 & 9089.82 & 10440.54\\
 \hline
        \end{tabular}
    \label{tab_ess}
    \end{center}
\end{table}

\subsubsection{Autocorrelation in MCMC methods}

The family of MCMC algorithms that we consider makes an essential assumption - every new sample accepted in the chain depends only on the last sample added to the chain. Thus autocorrelation inherently exists in the produced samples. 

\begin{figure}[htbp]
\centerline{\includegraphics[width=\columnwidth]{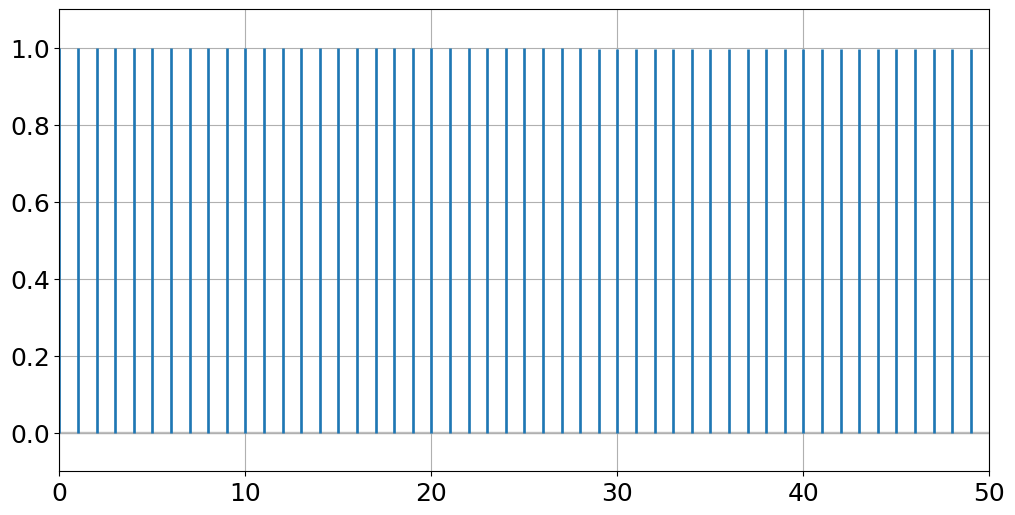}}
\caption{Autocorrelation plot from Metropolis-Hastings for the slope parameter \(k\) in the Linear Trend component}
\label{fig_ac_mh}
\end{figure}

We show the MH plot of the slope of the linear trend in Fig. \ref{fig_ac_mh}, noting that the DMZ plot looks similar. The other parameters behave the same, with a remark that the parameters that define the change points and seasonalities have slightly better autocorrelation plots.

\begin{figure}[htbp]
\centerline{\includegraphics[width=\columnwidth]{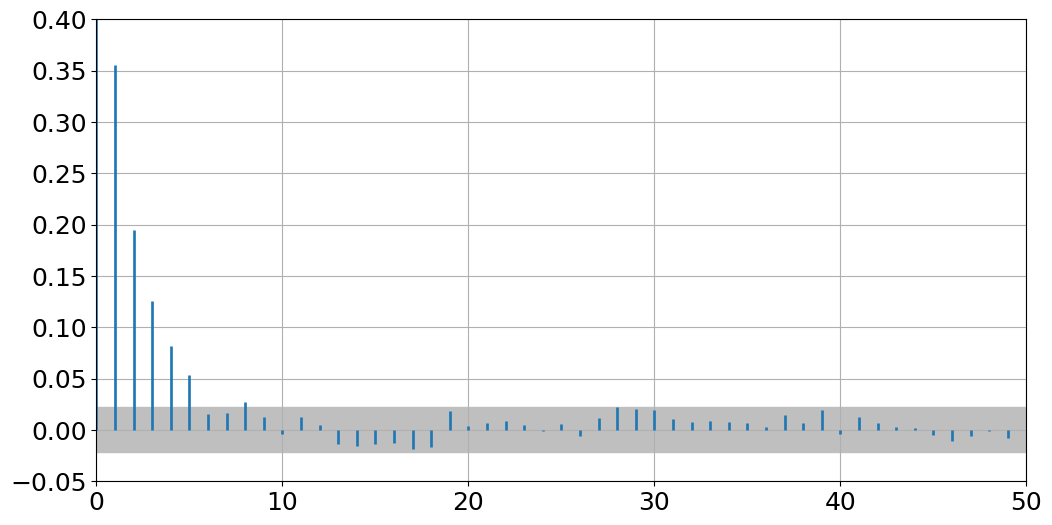}}
\caption{Autocorrelation plot from NUTS for the slope parameter \(k\) in the Linear Trend component}
\label{fig_ac_nuts}
\end{figure}

The autocorrelation plots of the posterior samples confirm what we found out from the ESS comparison. MH and DMZ generate autocorrelated chains that do not explore the posterior efficiently. Even with a lag of 50, the sample from MH shows a strong autocorrelation. NUTS, on the other hand, starts off with a much smaller autocorrelation for lags smaller than 5, and the autocorrelation gradually vanishes, as can be seen in Fig. \ref{fig_ac_nuts}.

The presented diagnostics show that NUTS converges well for all parameters.
To improve the MH and DMZ methods, we may need to draw more samples, run more chains, tune the sampler or adjust the proposal distribution or step sizes. But even if we generate a long chain of samples, still the amount of independent information may be low, due to the correlation of the samples.

\subsection{VI convergence}

\begin{figure}[H]
\centerline{\includegraphics[width=\columnwidth]{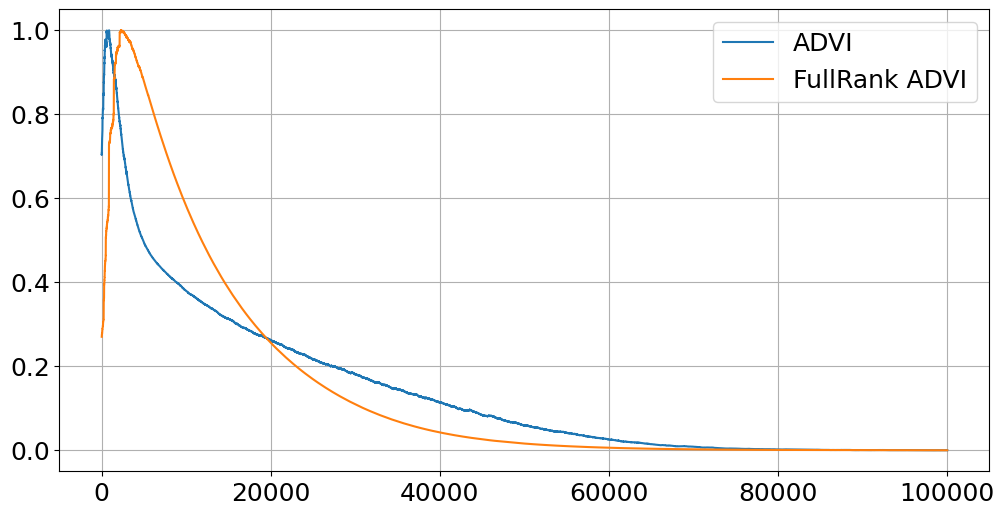}}
\caption{Negative ELBO through iterations}
\label{fig_vi_elbo}
\end{figure}

In the VI methods, the Evidence Lower Bound (ELBO) is the objective function that is maximized in order to find the best approximation to the true posterior. When the loss function (negative ELBO) has reached a steady state, it indicates that the variational inference procedure has effectively converged. However, this does not guarantee a high-quality posterior approximation. We need to monitor the ELBO curve and perform other diagnostic checks. To determine when the negative ELBO reaches a plateau, we plot the negative ELBO for both ADVI and FR ADVI through the iterations in Fig. \ref{fig_vi_elbo}.

To reduce noise, the plots are smoothed using exponential smoothing with a coefficent of \(10^{-3}\) for ADVI and \(10^{-4}\) for FR ADVI. Additionally, they are scaled in the \([0, 1]\) range because their ELBOs differ in absolute value, making it more difficult to notice the convergence. We observe that convergence is achieved around the \(80000\)-th iteration for both approaches.

\subsection{Forecasting metrics}

Table \ref{tab_mcmc_metrics} presents the standard forecasting metrics for the different MCMC inference techniques after drawing all samples and at their point of convergence. 

\begin{table}[H]
    \caption{Comparing MCMC forecasting metrics after drawing all samples and at point of convergence}
    \begin{center}
        \begin{tabular}{|c|c|c|c|c|c|}
        \hline
        \textbf{Method} & \textbf{MSE} & \textbf{RMSE} & \textbf{MAE} & \textbf{MAPE}\\
        \hline
        \textit{MH} & 0.6886 & 0.8298 & 0.6903 & 0.0862 \\
        \hline
        \textit{DMZ} & 0.5789 & 0.7609 & 0.6180 & 0.0770  \\
        \hline
        \textit{NUTS} & 0.6720 & 0.8198 & 0.6792 & 0.0848  \\
        \hline
        \hline
        \textit{NUTS at convergence} & 0.6584 & 0.8114 & 0.6698  & 0.0836\\
        \hline
        \end{tabular}
    \label{tab_mcmc_metrics}
    \end{center}
\end{table}

Since MH and DMZ never converge, we only show the metrics at the point of convergence for NUTS. We observe that there is no significant difference at the point of convergence and after drawing 2000 samples for NUTS.

Interestingly enough, eventough the DMZ method never converges, it still generates forecasts with the best quality.

Table \ref{tab_vi_metrics} presents the standard forecasting metrics for the different VI inference techniques after 100000 iterations and at their point of convergence. Increasing the number of iterations in the VI methods improves the metrics after reaching the point of convergence. ADVI generates forecasts with the highest quality.

\begin{table}[H]
    \caption{Comparing VI forecasting metrics}
    \begin{center}
        \begin{tabular}{|c|c|c|c|c|c|}
        \hline
        \textbf{Method} & \textbf{MSE} & \textbf{RMSE} & \textbf{MAE} & \textbf{MAPE}  \\
        \hline
        \textit{FR ADVI} & 0.7966 & 0.8925 & 0.7572 & 0.0947  \\
        \hline
        \textit{ADVI} & 0.7092 & 0.8421 & 0.7062 & 0.0882  \\
        \hline
        \hline
        \textit{FR ADVI at conv.} & 1.1355 & 1.0656 & 0.9325 & 0.1168  \\
        \hline
        \textit{ADVI at convergence} & 1.2841 & 1.1332 & 1.0093 & 0.1265  \\
        \hline
        \end{tabular}
    \label{tab_vi_metrics}
    \end{center}
\end{table}

Finally, in Table \ref{tab_map_metrics}, we present the metrics obtained with the MAP method.

\begin{table}[H]
    \caption{Metrics from the MAP method}
    \begin{center}
        \begin{tabular}{|c|c|c|c|c|c|}
        \hline
        \textbf{Method} & \textbf{MSE} & \textbf{RMSE} & \textbf{MAE} & \textbf{MAPE} \\
        \hline
        \textit{MAP} & 0.6915 & 0.8316 & 0.6925 & 0.0865 \\
        \hline
        \end{tabular}
    \label{tab_map_metrics}
    \end{center}
\end{table}

\subsection{Plot of the posterior samples}

Next, we see what the histograms of the samples drawn from the posteriors with different techniques look like. We are looking at the posterior samples drawn from the slope parameter.

\begin{figure}[htbp]
\centerline{\includegraphics[width=\columnwidth]{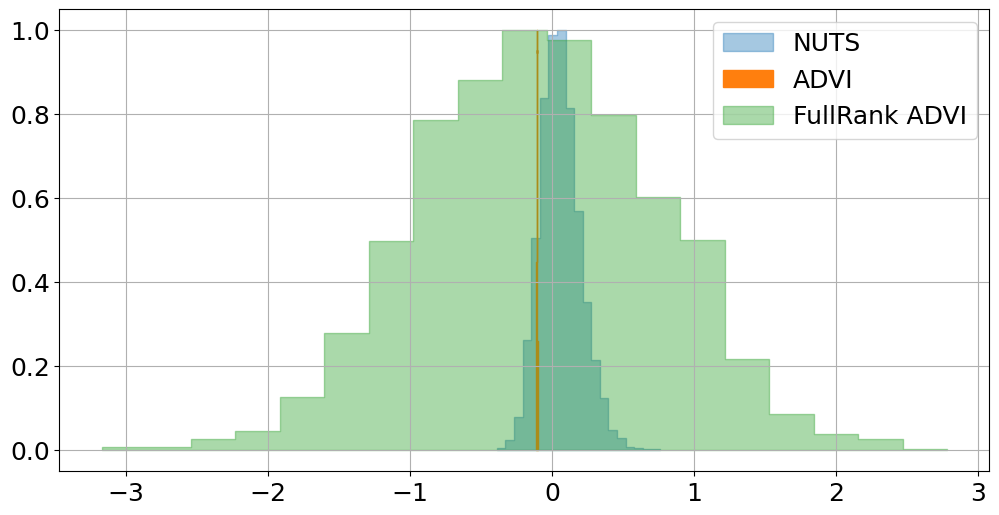}}
\caption{NUTS vs ADVI and FullRank ADVI for the slope parameter \(k\) in the Linear Trend component}
\label{fig_post_mcmc_advi}
\end{figure}

Fig. \ref{fig_post_mcmc_advi} shows that the sample drawn from the FullRank ADVI approximation is less "certain" about the value of the slope, i.e. it has a higher standard deviation than the sample drawn from the true posterior via the NUTS algorithm. On the other hand, the sample drawn from the ADVI approximation is overconfident about the value of the slope, i.e. it has a lower standard deviation than the sample drawn from the true posterior via the NUTS algorithm.

\begin{figure}[htbp]
\centerline{\includegraphics[width=\columnwidth]{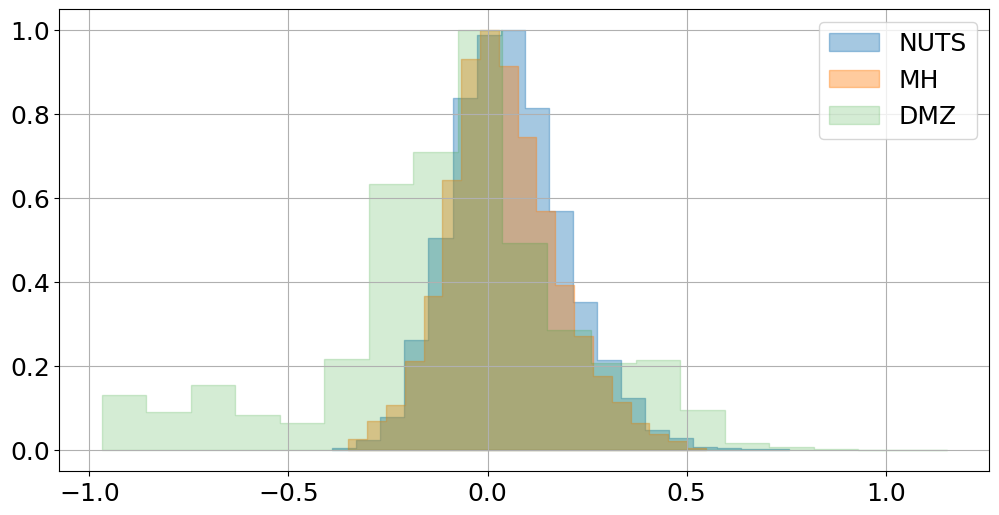}}
\caption{NUTS vs MH vs DMZ for the slope parameter \(k\) in the Linear Trend component}
\label{fig_post_nuts_mh}
\end{figure}

Fig. \ref{fig_post_nuts_mh} shows that the sample drawn via the MH method is similar with the sample drawn via the NUTS method, but the DMZ method produces a slightly less confident sample.

\subsection{Summarized results}

Finally, we summarize the performance of the various Bayesian inference methods. Table \ref{tab_sumarized_mcmc} shows the number of samples drawn and the time required for those samples to be drawn for each MCMC. We also present the number of samples and the time required for these methods to converge, noting that only NUTS successfully converges. The last column shows the estimated number of effective samples drawn per second - effective sampling rate is a measure of computational efficiency.

\begin{table}[H]
    \caption{Summarized results for the MCMC methods}
    \begin{center}
        \begin{tabular}{|c|c|c|c|}
        \hline
        & \textbf{MH} & \textbf{DMZ} & \textbf{NUTS} \\
        \hline
        \textit{\# Samples} & 1000000 & 1000000 & 2000 \\
 \hline
\textit{\# Samples to convergance} & / & / & 550\\
 \hline
\textit{Time (s)} & 7141.66 & 387.13 & 193.77 \\
 \hline
\textit{Time (s) to convergance} & / & / & 87.04 \\
 \hline
 \textit{ESS per s} & 150.61 & 1.14 & 169.69 \\
 \hline
        \end{tabular}
    \label{tab_sumarized_mcmc}
    \end{center}
\end{table}

NUTS achieves convergence i.e. achieves an R-hat diagnostic statistic less than 1.01 and achives an ESS larger than 400, after only drawing 550 samples. This means that NUTS is very effective, with at least 17\% of the samples drawn being effective samples.

We present a similar summarization in Table \ref{tab_sumarized_vi} for the VI methods, only this time we focus on the time required for the total number of iterations of the models and the time required for the models to converge.

\begin{table}[H]
    \caption{Summarized results for the VI methods}
    \begin{center}
        \begin{tabular}{|c|c|c|}
        \hline
        & \textbf{FR ADVI} & \textbf{ADVI}\\
        \hline
        \textit{\# Iterations} & 100000 & 100000 \\
 \hline
\textit{\# Iterations to convergance} & $\approx$80000 & $\approx$80000\\
 \hline
\textit{Time (s)} & 22.97 & 18.25 \\
 \hline
\textit{Time (s) to convergance} & 18.85 & 15.06 \\
 \hline
        \end{tabular}
    \label{tab_sumarized_vi}
    \end{center}
\end{table}

\section{Conclusion}\label{conclusion}

In this paper we analised  three major Bayesian inference classes: MAP estimation, full MCMC posterior sampling, and VI on a time-series forecasting problem using the Facebook Prophet model. 

MAP estimation consistently delivered high-quality forecasts and benefits from fast execution times. Therefore, MAP is particularly suitable in scenarios where accurately capturing posterior shape and uncertainty is less critical. However, it is important to note that MAP estimates the mode of the posterior distribution, which may differ from the mean usually used in forecasting via MCMC/VI.

MCMC sampling generally provided higher-quality posterior samples compared to VI approaches. This is not surprising since we had to draw large number of samples for MH and DMZ and these methods explore the true posterior. Still, there are some convergence issues that have to be addressed. 

Given the computational cost associated with full MCMC sampling, we may consider using ADVI or FullRank ADVI. The forecasting metrics achieved with VI methods are comparable with those of the full MCMC sampling algorithms, but the execution time is reduced  significantly. However the FullRank ADVI overestimated and ADVI underestimated posterior uncertainty. This could be due to algorithmic limitations and should be further addressed. Thus NUTS maybe be the best choice for our problem, given the current settings. Its time to convergence  is higher compared to VI methods, but not as high as the other MCMC methods. Additionally NUTS showed overall good results for all conducted analyses. 

We give a slightly different interpretation of the results that offers some intuition on when to use each technique. We already stated that the MAP estimation is equal to performing MLE with regularization. We can go even further, and state that in the case of Laplace or Normal priors, MAP estimation is equal to performing MLE with L1 or L2 regulatization, respectively (for a Laplace prior with scale \(b\), the regularization strength is \(\lambda=\frac{1}{b}\), and for a Normal prior with scale \(\sigma\), the regularization strength is \(\lambda = \frac{1}{2\sigma^2}\)).

Then, if MCMC is not fast enough, VI can be used to approximate the uncertainty in these parameters with Laplace and Normal priors. This uncertainty approximation can later be used to determine the regularization strength in both Frequentist and Bayesian frameworks. 

\section{Future work}

As far as the research of various Bayesian inference techniques goes, in the future we will be looking into GPU acceleration to speed up the full MCMC sampling. Packages like JAX\cite{jax2018github} and NumPyro\cite{bingham2019pyro} offer GPU acceleration, and they integrate well with PyMC, showing improvements in execution time on large datasets with large models, when the computation of the likelihood is expensive.

However, we need to note that every MCMC algorithm is inherently sequential because of the underlying Markov assumption. In other words, to produce the \(i\)-th sample from the posterior, one must produce the \((i-1)\)-th sample first and there is no approach that can produce these samples in parallel. Thus, GPU acceleration can only speed up the computation of the likelihood. Even then, if the likelihood is simple, the model does not involve a lot of parameters or the dataset is small, GPU acceleration may end up adding a computational overhead that will overcome the benefits of fast matrix operations, leading to an overall slower sampling approach than a CPU based one.

VI, on the other hand, does not have this sort of limitation. That means that most of our future research will consist of optimizing the VI approximation process by utilizing GPU acceleration. One additional thing worth investigating is the overestimation and underestimation of the posterior uncertainty with FullRank ADVI and ADVI respectively.

Finally, we want to further investigate the usage of Bayesian inference to determine the regularization strength of various parameters, as was mentioned in section \ref{conclusion}.

\bibliographystyle{IEEEtran}
\bibliography{IEEEabrv,refs}

\end{document}